\documentclass[journal]{IEEEtran}
\usepackage{cite}
\usepackage{amsmath,amssymb,amsfonts}
\usepackage{graphicx}
\usepackage{textcomp}
\usepackage{epstopdf}
\usepackage{lineno}
\usepackage{algpseudocode}
\usepackage{algorithmicx,algorithm}
\usepackage{subfigure}
\usepackage{booktabs}
\usepackage{multirow}
\usepackage{makecell}
\usepackage[T1]{fontenc}
\usepackage[utf8]{inputenc}
\usepackage{authblk}
\usepackage{amsthm}
\usepackage{bm}
\usepackage{hyperref}
\usepackage{bbding}
\usepackage[table]{xcolor}
\usepackage{colortbl}
\begin{document}
%Hybrid Federated and Split

\title{Distributed Machine Learning in D2D-Enabled Heterogeneous Networks: Architectures, Performance, and Open Challenges}
\author{Zhipeng~Cheng,~\IEEEmembership{Member,~IEEE,} Xuwei Fan, Minghui Liwang,~\IEEEmembership{Member,~IEEE}, Ning Chen, Xiaoyu Xia,~\IEEEmembership{Member,~IEEE} and Xianbin Wang,~\IEEEmembership{Fellow,~IEEE}

\thanks{Zhipeng Cheng is with Soochow University, China; Xuwei Fan, Minghui Liwang (Corresponding author) and Ning Chen are with Xiamen university, China; Xiaoyu Xia is with RMIT University, Australia; Xianbin Wang is with Western University, Canada.}
}

\maketitle
\thispagestyle{empty}
\pagestyle{empty}
\begin{abstract}
The ever-growing concerns regarding data privacy have led to a paradigm shift in machine learning (ML) architectures from centralized to distributed approaches, giving rise to federated learning (FL) and split learning (SL) as the two predominant privacy-preserving ML mechanisms. However, implementing FL or SL in device-to-device (D2D)-enabled heterogeneous networks with diverse clients presents substantial challenges, including architecture scalability and prolonged training delays. To address these challenges, this article introduces two innovative hybrid distributed ML architectures, namely, hybrid split FL (HSFL) and hybrid federated SL (HFSL). Such architectures combine the strengths of both FL and SL in D2D-enabled heterogeneous wireless networks. We provide a comprehensive analysis of the performance and advantages of HSFL and HFSL, while also highlighting open challenges for future exploration. We support our proposals with preliminary simulations using three datasets in non-independent and non-identically distributed settings, demonstrating the feasibility of our architectures. Our simulations reveal notable reductions in communication/computation costs and training delays as compared to conventional FL and SL.
\end{abstract}

\IEEEpeerreviewmaketitle
\section{Introduction}
The past decade has witnessed the expeditious evolution of communication and computing technologies, and their proliferation in many emerging fields such as the Internet of Vehicles and E-health, where massive amount of data is generated, exchanged and utilized. Such developments have brought both technical challenges and great opportunities for a wide range of machine learning (ML)-based applications, since ML holds considerable promise to fast decisions and inferences without human intervention \cite{MyNetwork, SL-Healthcare-2018}. Besides, device-to-device  (D2D) communications-enabled multi-layer heterogeneous wireless networks have become one of significant components of 5G/6G networks, where the complicated network topologies could impose great challenges to the implementations of ML-based applications\cite{FogLearning}.

Securing abundant training data and computation resources represents the fundamental requirement of ML. Traditional ML is generally centralized where massive data are collected and transmitted from local devices to centralized data centers associated with remote cloud servers. Disregards its advantages such as high accuracy and efficiency, centralized ML confronts the following deficiencies:
\begin{itemize}
  \item When a large volume of training data has to be transmitted over dynamic wireless networks, excessive transmission delay as well as possible increased power consumption for local devices can be incurred.
  \item Centralized ML is often less effective for rapid model deployment. Meanwhile, it suffers from unsatisfying scalability in large-scale networks, especially when the model requires frequent retraining.
  \item Centralized ML is difficult to reserve data privacy, as many applications may involve the uploading of large amount of private information. For instance, regarding pathological pictures in E-health\cite{SL-Healthcare-2018}, most local devices (e.g., patients) are generally unwilling to provide privacy-sensitive data due to ever-growing privacy concerns, which can result in a dilemma between model training and privacy protection.
\end{itemize}
% 271
\subsection{Preliminaries of FL and SL}
\begin{figure}[t]
	\centering
	\includegraphics[width=3.4in]{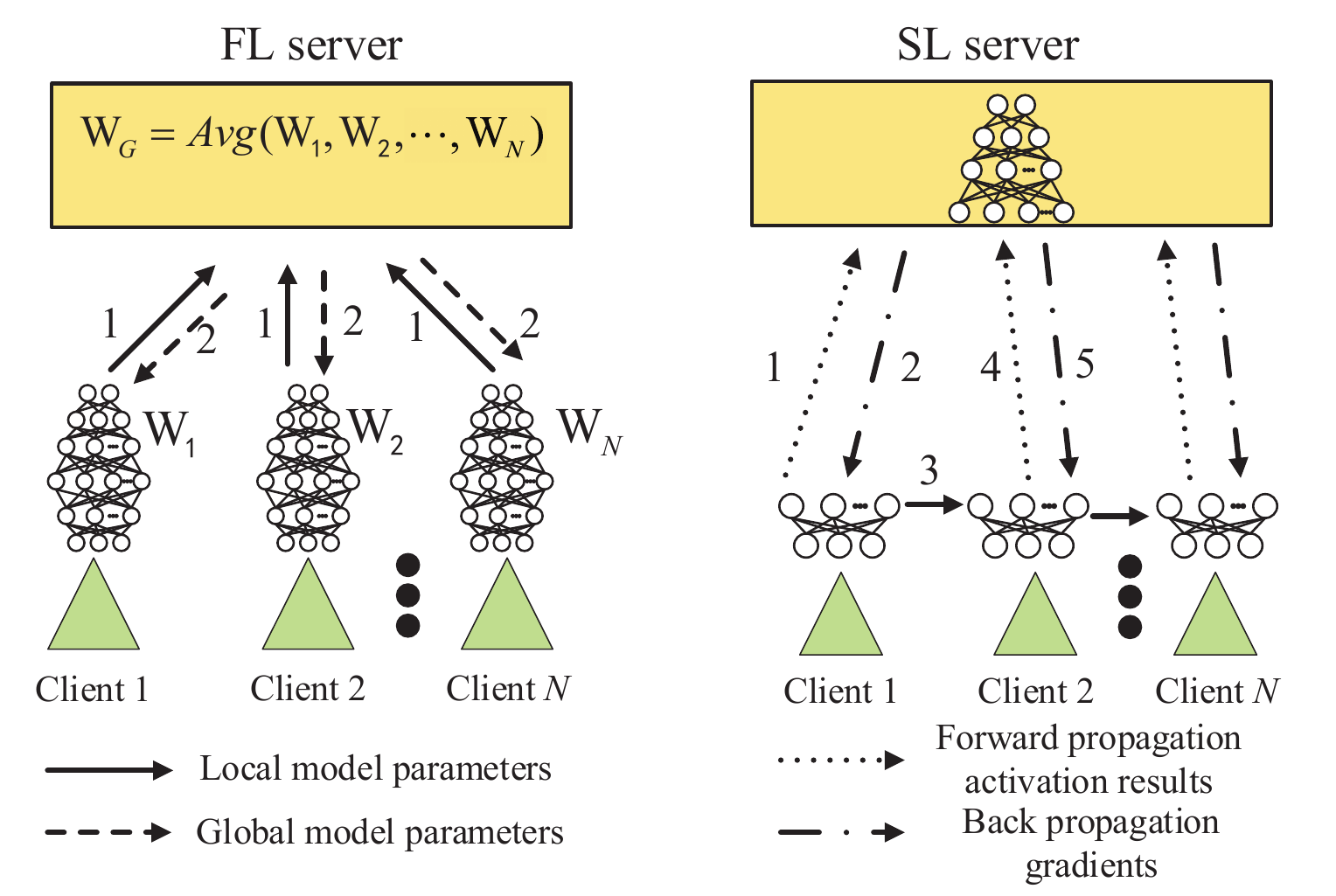}
	%via \DeclareGraphicsExtensions.
	\caption{An illustration of the learning architectures of federated learning (the left side) and split learning (the right side).}
	\label{fig1}
\end{figure}

To reconcile the demand for ML model training and privacy protection, a straight idea is to conduct the model training by distributed data owners to avoid sharing raw data, which facilitates distributed ML architectures. Specifically, federated learning (FL) \cite{FL-2016} and split learning (SL) \cite{SL-2018} represent two bright peals. Specifically, these two learning mechanisms implement distributed ML from different perspectives, while their representative learning architectures are depicted in Fig. \ref{fig1}.

\textbf{Training process of FL}: A typical FL scheme engages a set of smart devices termed as \textit{clients} to participate in the iterative model training. At the beginning of each iteration, each client receives a global model from a parameter server, then conducts local training to update the model by performing stochastic gradient descent on its local training data. After the completion of local training, all clients involved in FL upload the trained model parameters to the FL server in parallel (step 1). The FL server next aggregates (e.g., FedAvg $Avg(\cdot)$ ) the overall received model parameters into a new global model, which will be broadcasted to clients (step 2) for the following training round. Specifically, each client only exchanges the model parameters with the FL server, preventing privacy disclosure to some extent.

\textbf{Training process of SL}: To achieve SL training, an ML neural network is firstly split into two subnetworks via splitting in the middle layer (i.e., a split layer) of the network. Generally, the subnetwork associated with input layer is deployed on the clients' side, and the one related with output layer will be deployed at an SL server. In each iteration, a client starts training by performing forward propagation; and then, the activation data, i.e., the output of split layer of the client (with label data), will be transmitted to the SL server (step 1). Upon receiving the activation data, the SL server proceeds with forward propagation within its subnetwork to derive output results, and subsequently initiates the back propagation process, which involves the computation of the loss by comparing the output results with the provided label data\cite{SL-2018}. Similarly, the output of cut layer can be transmitted back to the client for subsequent back propagation (step 2). Finally, subnetworks on both the client and server can be updated respectively. By offloading the model parameter of the current client to the next client (step 3) and repeating the above steps, the model can be trained sequentially over multiple clients.
% 419

\subsection{Motivations and Contributions}
Interestingly, FL and SL share some common advantages, such as communication/computation cost reduction, and privacy preservation. However, the implementations of FL and SL in heterogeneous wireless networks can also encounter non-negligible challenges:
\begin{itemize}
  \item \textbf{Network heterogeneity}: The SL server can only interact with the clients sequentially, leading to a low level of parallelization and may hinder the corresponding convergence speed. For FL, frequent model reporting from clients to the server over unstable wireless links usually result in model performance degradation due to transmission failures. Besides, since the implementation of conventional FL and SL generally relies on a star topology, where a central server coordinates the model update of all the clients. Thus, FL and SL may suffer from unsatisfying scalability and performance degradation in large-scale D2D-enabled heterogeneous networks due to single point failures\cite{FogLearning}. For example, under a space-air-ground-ocean integrated 6G network architecture, satellites, unmanned aerial vehicles, smart vehicles, and underwater unmanned vehicles can be the clients, where frequent D2D communications among them can further complicate the network topology (e.g., hierarchical tree topology). Such disadvantages advocate combining both FL  and SL architectures, to guarantee the training performance in heterogeneous wireless networks.

  \item \textbf{Client heterogeneity}: Clients generally have heterogeneous computation/communication/energy capabilities. Each client in FL requires more computation/energy resources to support model training. Besides, since the stability of wireless communications are essential to guarantee successful model transmissions (e.g., the size of a large complete model can reach 1 GB\cite{BAcombo}), FL prefers clients with sufficient computation/energy/communication resources. Differently, SL usually supports clients with constrained on-board computation/energy resources since each client only has to train a partial model, which, however, can incur heavy communication overhead. More importantly, imbalanced and non-independent and identically distributed (non-IID) data of clients can leave heavy impacts on training performance. To this end, exploring the combination of FL and SL architectures to make full use of clients' heterogeneous capabilities and resources presents another major significance.
  \item \textbf{Optimization target heterogeneity}: Typically, model test accuracy and convergence speed represent the key optimization targets of distributed ML\cite{FogLearning}. Besides, when distributed ML is implemented in heterogeneous networks, common indicators such as datarate, throughput, delay, and energy consumption also represent major concerns of ML, which complicates problems such as client scheduling and resource allocation\cite{MultiFL-TMC}. Thus, it is significant to improve distributed ML architectures according to the characteristics of heterogeneous networks while realizing multiple targets optimization.
\end{itemize}

Given the above discussed challenges and limitations of conventional FL and SL in heterogeneous networks, this article is motivated to investigate two comprehensive architectures upon considering the combination of FL and SL. Our main contributions are highlighted below:
\begin{itemize}
  \item We first introduce a hybrid split FL (HSFL) architecture by integrating the split architecture into FL. Then, we are interested in designing a hybrid federated SL (HFSL) architecture, which unifies federated architecture with SL. key topics such as the advantages and performance comparisons of the two architectures are discussed.

   \item  Several open challenges are comprehensively discussed to identify the future research directions of our proposed architectures for future implementations.

  \item Preliminary simulations are conducted to verify the feasibility of our proposed architectures on three datasets, under highly non-IID data settings.
\end{itemize}
% 527
\begin{figure*}[t]
    \centering
    \subfigure[] {\includegraphics[width=3.1in,angle=0]{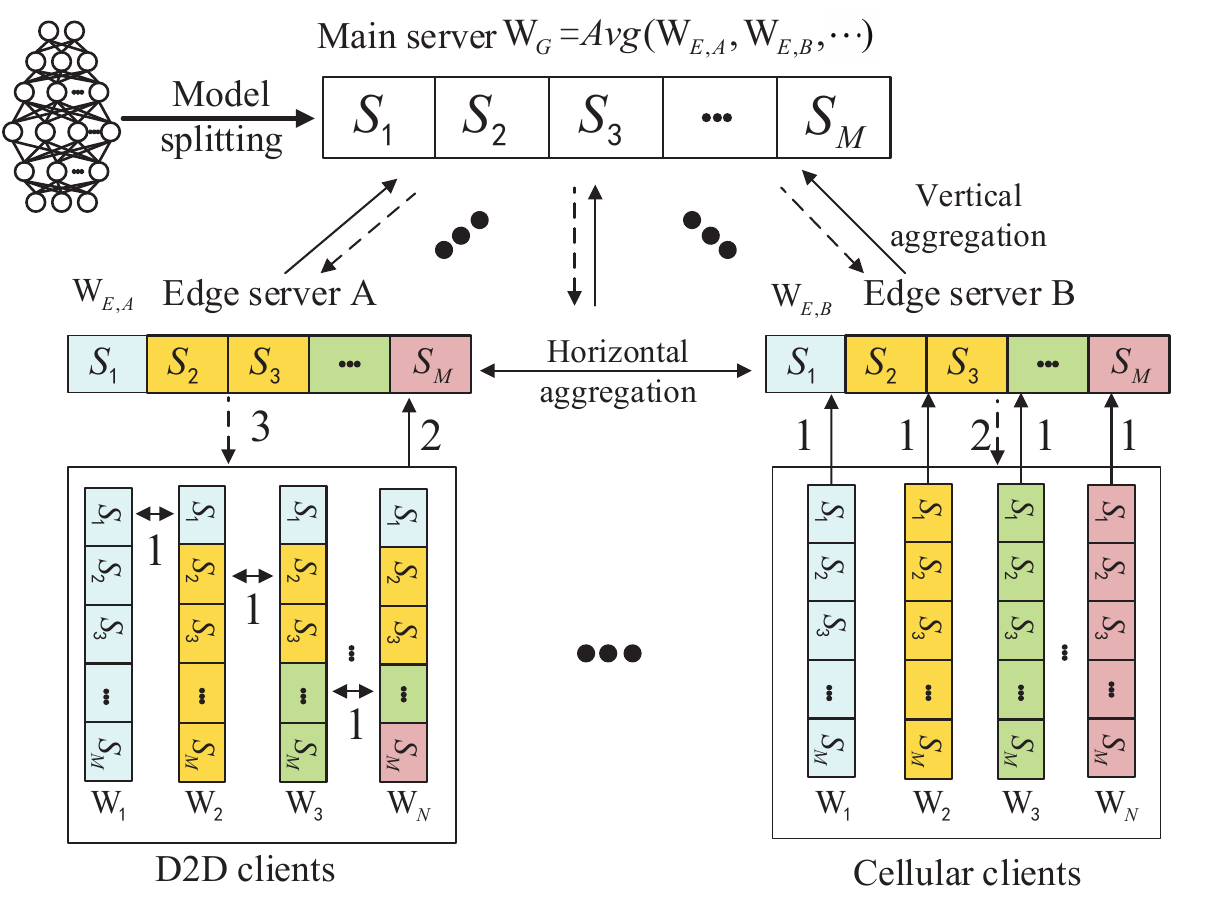}}
    \subfigure[] {\includegraphics[width=3.6in,angle=0]{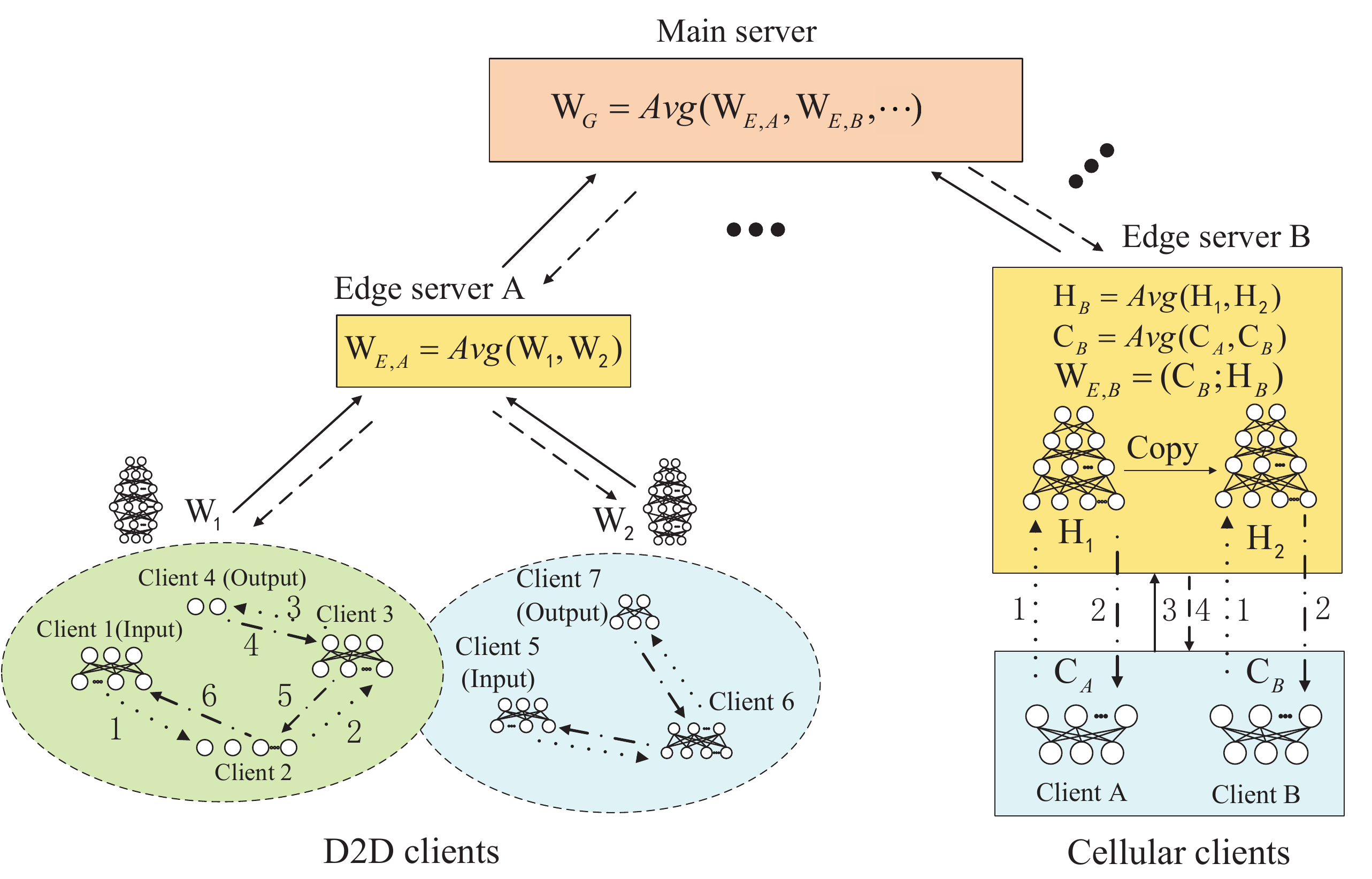}}
    \caption{The proposed hybrid ML architectures in a multi-layer D2D-aided heterogeneous network: a) a schematic of the proposed HSFL architecture; b) a schematic of the proposed HFSL architecture.}
    \label{fig2}
\end{figure*}

%\begin{figure*}[!htbp]
%	\centering
%	\includegraphics[width=4in]{Fig2.eps}
%	%via \DeclareGraphicsExtensions.
%	\caption{ .}
%	\label{fig2}
%\end{figure*}
%
%\begin{figure}[!htbp]
%	\centering
%	\includegraphics[width=3in]{Fig3.eps}
%	%via \DeclareGraphicsExtensions.
%	\caption{ .}
%	\label{fig3}
%\end{figure}

\section{Investigations}
Some early efforts have been devoted to exploring the combination of FL and SL, as well as the corresponding performance improvement. For HSFL, \cite{BAcombo} proposed a decentralized FL mechanism (i.e., gossip learning) based on FL model splitting in a D2D network, where each FL client only transmits model segmentations to neighboring clients. However, this work only considers the D2D network. For HFSL, \cite{PSL-2020} proposed a parallel SL framework, where all the clients' subnetworks are synchronized. In each training round, every client sends the whole gradients back to the server, while then the server averages the gradients and transmits them back to the clients. Although such a framework enables SL with parallelism during clients' model update process, it relies on a single server and thus results in poor scalability, especially in large-scale networks. \cite{J-SplitFed-20} studied a SplitFed framework where the clients' model parameters are also averaged with a dedicated server. Besides, the subnetwork at the server side is updated by averaging the gradients of each client. However, unsatisfying scalability represents one of the key drawbacks of SplitFed, upon considering a raising number of clients\cite{C-CLOUD-FSL-21}. Therefore, \cite{C-CLOUD-FSL-21} decided to deploy edge servers as coadjutants to alleviate the communication and computation load of the SL server; then each edge server can interact with one or multiple clients for gradients exchange; while the SL server can further calculate the averaged gradients and update the subnetworks at edge servers. \cite{J-TC-SFLG-22} put forward similar ideas by deploying multiple FL servers to handle groups of clients. Although the above discussed works have made certain efforts, none of them have comprehensively analyzed the implementations on integrating FL and SL in D2D-enabled heterogeneous networks.

\section{Architectures}
\subsection{Architecture of HSFL}
According to previous discussions, FL clients may undergo heavy communication costs since each of them has to transmit a complete model to the server, especially when facing with large size models over unstable wireless links. Besides, in typical FL, the model of a client is no longer useful in updating the global mode when confronting transmission failures, e.g., only partial of the corresponding model has been successfully transmitted to the server. Promoted by the principle of SL and to overcome the drawbacks of FL, we delve into splitting the model from a different perspective, namely, the number of model parameters. Accordingly, a comprehensive HSFL architecture is proposed as inspired by \cite{BAcombo}, in a multi-layer heterogeneous wireless networks (as depicted in Fig. \ref{fig2}(a)), consisting of D2D clients, cellular clients, edge servers, and a main server. Key modules of HSFL are detailed below.

\textbf{Model splitting:} The model is firstly split into $M$ ($\{S_{1},\dots,S_{M}\}$) segments with equal data size where each segment is identified by a unique identification number. $M$ represents a hyperparameter that can be different for various FL models. More importantly, a larger $M$ can bring a smaller model granularity, while a higher transmission efficiency could be reached. Considering different communication conditions, an appropriate value of $M$ can ensure a good trade-off between transmission capacity and communication efficiency. Although each client can set $M$ by itself theoretically, to facilitate model aggregation/storage, all clients will use the same value of $M$.

\textbf{Model transmission and aggregation at clients:} Each client first evaluates the wireless channel quality and transmission capacity, to determine the number of segments that can be transmitted successfully. Any specific segments for transmission can be randomly chosen or specified by the receiver, while different clients can transmit the same segments. Specifically, for D2D clients, suppose that each client can communicate with its neighboring clients within one hop, it will send/receive at least one segment to/from each neighbor. Two paradigms associated with model transmission and model aggregation are applied for D2D clients and cellular clients, respectively, as given in Fig. \ref{fig2}(a). On the right side, cellular clients can transmit model segments to the edge server in parallel, for example, client 1 sends segment 1 to the edge server while client 2 sends segments 2 and 3. On the left side of Fig. \ref{fig2}(a), D2D clients transmit model segments to their neighboring clients sequentially in a decentralized manner, while the last D2D client can send the aggregated model to the edge server. Specifically, edge servers or D2D clients proceed with segment-wise model aggregation, where the model segments are aggregated individually. For example, edge servers aggregate segment 1 of $W_{E,B}$ by averaging all the received segments 1.

\textbf{Horizontal/Vertical model aggregation at edge servers:} Both vertical aggregation and horizontal aggregation are considered in HSFL to improve communication efficiency. For vertical model aggregation, the model transmission and aggregation can be repeated for multiple rounds to obtain multiple model replicas, which thus reduces the communication cost between edge servers and main server\cite{FogLearning}. Then, the model will be transmitted to the main server for a wide-range global aggregation. Horizontal aggregation among edge servers can be seen as a unique form of D2D communication at the edge server level. This approach effectively diminishes the overall model size transmitted to the main server, thereby reducing communication costs. Additionally, horizontal aggregation facilitates model parameter sharing, resulting in a significant acceleration of local model training and the mitigation of the impact of non-IID data distributions among clients.

\subsection{Architecture of HFSL}
Fig. \ref{fig2}(b) illustrates the proposed HFSL architecture, where the main principle is to parallelize SL training and average the model weights over multiple clients by applying FL. Similar to HSFL, both D2D clients and cellular clients are considered.

When HFSL is implemented over multiple D2D clients, clients can be clustered into several clusters, e.g., based on their communication/computation capabilities. Then, the ML model is split into multiple subnetworks and distributed to the clients within each cluster. As shown on the left side of Fig. \ref{fig2}(b), the model is split into 4, and 3 subnetworks for two clusters, respectively. Then, the forward propagation starts at the client (e.g., client 1) with the input layer while ending at the client (e.g., client 4) with the output layer. Next, the back propagation starts in reverse order. Thus, each D2D cluster can be regarded as a hyper FL client, training a complete model, i.e, ${W}_{1},{W}_{2}$.  The models can be transmitted by the clients to the edge server/neighboring cluster for averaging aggregation. In the next training round, the training starts with different clients under a new model splitting setting. Apparently, the training process is sequential within each D2D client cluster, and parallel over different clusters.

For cellular clients, we are motivated by \cite{PSL-2020, J-SplitFed-20,J-TC-SFLG-22, C-CLOUD-FSL-21}, and integrate them into the proposed HFSL. As shown by the right side of Fig. \ref{fig2}(b), the ML model is split into two subnetworks $C$ and $H$, where each client trains the same subnetwork $C$ in parallel while the subnetwork $H$ is deployed at edge server $B$. When multiple clients forward the activation results to the server in parallel, the edge server $B$ first copies the model to obtain multiple model copies to conduct forward propagation and back propagation for different clients in parallel. The number of model copies should be equal to the number of clients, e.g., two copies for clients $A$ and $B$. Then, the gradients will be sent back to the clients for updating the corresponding subnetwork parameters, $C_{A}$ and $C_{B}$; meanwhile the server can update parameters $H_{1}$ and $H_{2}$. Then, the clients can send the updated model parameters to the edge server in parallel. Finally, the edge server aggregates the model copies of edge server $B$ with $H_{B}$, and $C_{B}$ of the two clients. Among which, $C_{B}$ will be sent back to the clients, helping with the next training round. Notably, the number of cellular clients associated with each edge server should be optimized so that to alleviate the model storage cost of the edge server. Similarly, edge server $B$ can further report the complete model parameter $W_{E,B}$ to the main server for wide-range global aggregation. Note that horizontal model aggregation can also be done between the edge servers in HFSL, which is omitted here.

\section{Performance and Advantages}

\subsection{Performance Analysis}

Although both HSFL and HFSL represent concrete implementations combining FL and SL architectures, they exhibit both connections and distinctions. Firstly, the fundamental architecture of HSFL is FL, wherein all clients are responsible for training complete ML models and only need to transmit partial (or complete) model parameters. In contrast, HFSL's core architecture is SL, where each client exclusively trains a specific segment of the model. The federated architecture in HFSL is designed for parallel training and multi-layer model aggregation. Consequently, while HSFL and HFSL fundamentally represent two different architectures, they are both applicable in D2D-enabled heterogeneous networks. Generally, HSFL and HFSL are interconnected and can be harmoniously combined to create a complex yet effective architectural solution.

 To achieve better performance comparison regarding different learning architectures, we quantify the communication (comm.)/computation (comp.) cost of clients via a simple analytical analysis. Without losing generality, we mainly consider cellular clients, since it is challenging to compare the performance of different architectures under the same parameter settings for D2D clients. Besides, it is hard to find general settings for decentralized FL and SL. Assuming that there are $N$ clients, the total training data size is $D$ where each client has the same training data size $D/N$. The overall model data size is $|W|$. The model is split into two subnetworks for SL and HFSL, the size of the split layer is $b$, and the size of forward propagation or back propagation over the split layer can be calculated by $bD/N$\cite{MIT-FLSL-19}. The fraction of model size with clients is $\gamma$ and $(1-\gamma)$ with the server. For HSFL, the model is equally divided into $M$ segments, while each client transmits $m$ segments to the server at one training round. The computation cost, i.e., floating point of operations (FLOPs) of training a complete model is denoted by $F$; and the size of allocated computation load fraction at the client is $\lambda$. Thus, the total computation cost is $NF$ for FL and HSFL, and it is $N\lambda F$ for SL and HFSL. Correspondingly, HSFL reduces the total communication cost of FL from $2N|W|$ to $2Nm|W|/M$, while HFSL raises the communication cost of SL from $N(2bD/N+\gamma|W|)$ to $2N(bD/N+\gamma|W|)$, which, however, greatly reduces the training time through parallelization (as shown in the simulation).

%Besides, $T_{train}$ is the maximum time to train a complete model for any clients, and $T_{avg}$ is the required time for full model aggregation. For simplicity, suppose that the transmission datarate between any clients is $R_{c}$, while the $R_{e}$ is the upload/download transmission datarate between an edge server and clients.
\subsection{Key Advantages}
According to the above-discussions, the advantages of HSFL and HFSL are summarized as follows:
\begin{itemize}
  \item \textbf{Training cost reduction}: When compared to FL\/SL, both HSFL and HFSL can significantly reduce communication costs and training time. However, HSFL is more effective for training large-sized models, especially when dealing with a large number of clients, in contrast to HFSL.
  \item \textbf{Communication efficiency improvement}: Wireless spectrum resources can be efficiently reused through the in-band D2D communication mode for D2D and cellular clients, aided by an appropriate resource sharing scheme.
  \item \textbf{Data-efficient training}: HSFL can mitigate potential model transmission failures and make effective use of the local models trained by clients. Furthermore, HFSL can enable the participation of more clients with limited computational resources in each global training round, thanks to its parallelism capability. Additionally, the integration of D2D communications and a multi-layer hybrid network architecture significantly enhances server-client coverage and boosts the efficient utilization of training data.
  \item \textbf{Good adaptation in time-varying network topology}: Leveraging the hybrid network architecture, both HSFL and HFSL demonstrate strong adaptability to dynamic network conditions, including those resulting from client mobility, and they achieve excellent stability throughout the training process.
   \item \textbf{High privacy protection}: HSFL and HFSL empower clients to train or transmit partial models, thereby diminishing the likelihood of privacy disclosure resulting from malicious attacks or eavesdropping.
\end{itemize}

\

%\begin{figure*}[!htbp]
%    \centering
%    % \subfigure[] {\includegraphics[width=3in,angle=0]{fig3_reward.eps}}
%     %\subfigure[] {\includegraphics[width=3in,angle=0]{fig3_delay.eps}}
%      \subfigure[] {\includegraphics[width=2.4in]{Fig4.eps}}
%      \subfigure[] {\includegraphics[width=2.4in]{Fig5.eps}}
%      \subfigure[] {\includegraphics[width=2.4in]{Fig4.eps}}
%      %\subfigure[] {\includegraphics[width=3in,angle=0]{Fig5.eps}}
%    \caption{Test accuracy of CL, FL, SL, HSFL, and HFSL over training rounds with non-IID setting on three data sets: a) MNIST; b) Fashion-MNIST; c) CIFAR-10.}
%    \label{fig3}
%\end{figure*}
\begin{figure*}[htbp]
    \centering
    \subfigure[] {\includegraphics[width=2.3in,angle=0]{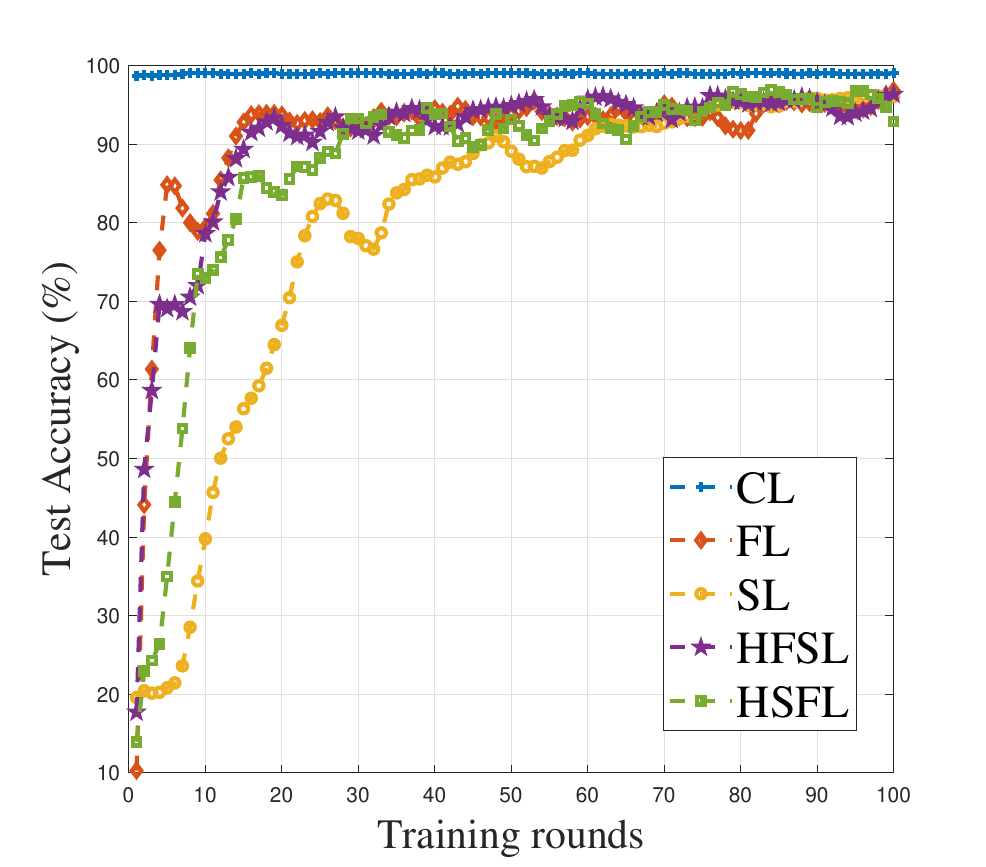}}
    \subfigure[] {\includegraphics[width=2.3in,angle=0]{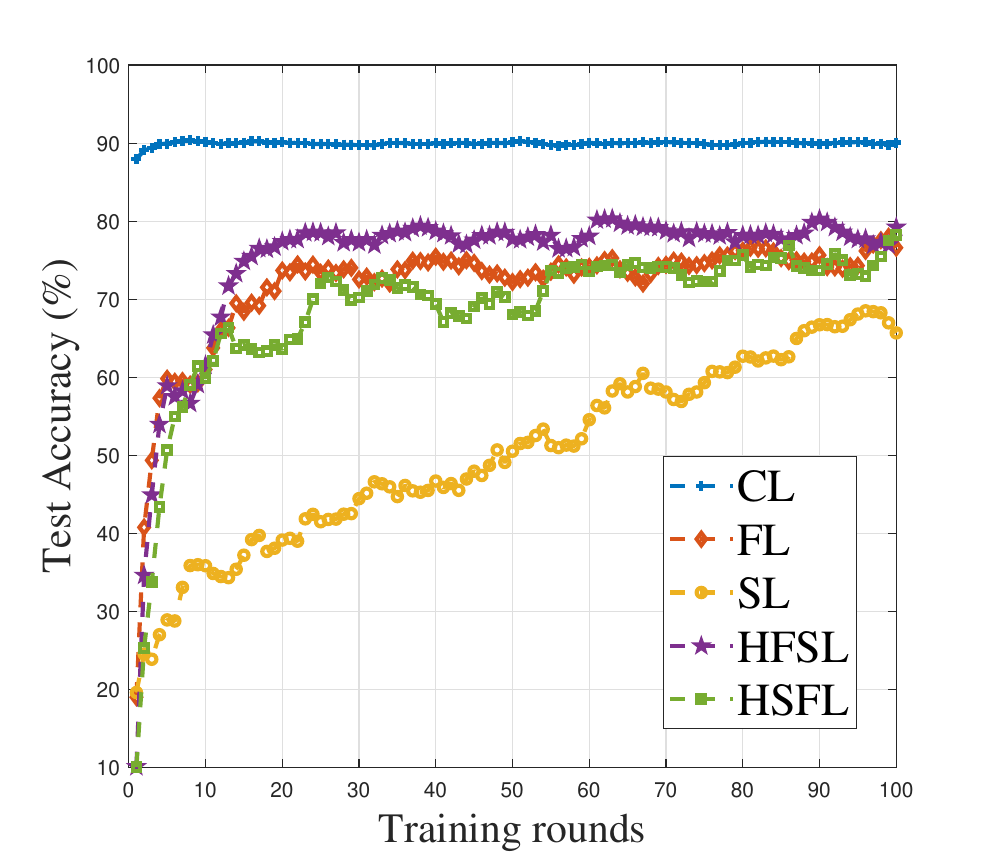}}
     \subfigure[] {\includegraphics[width=2.3in,angle=0]{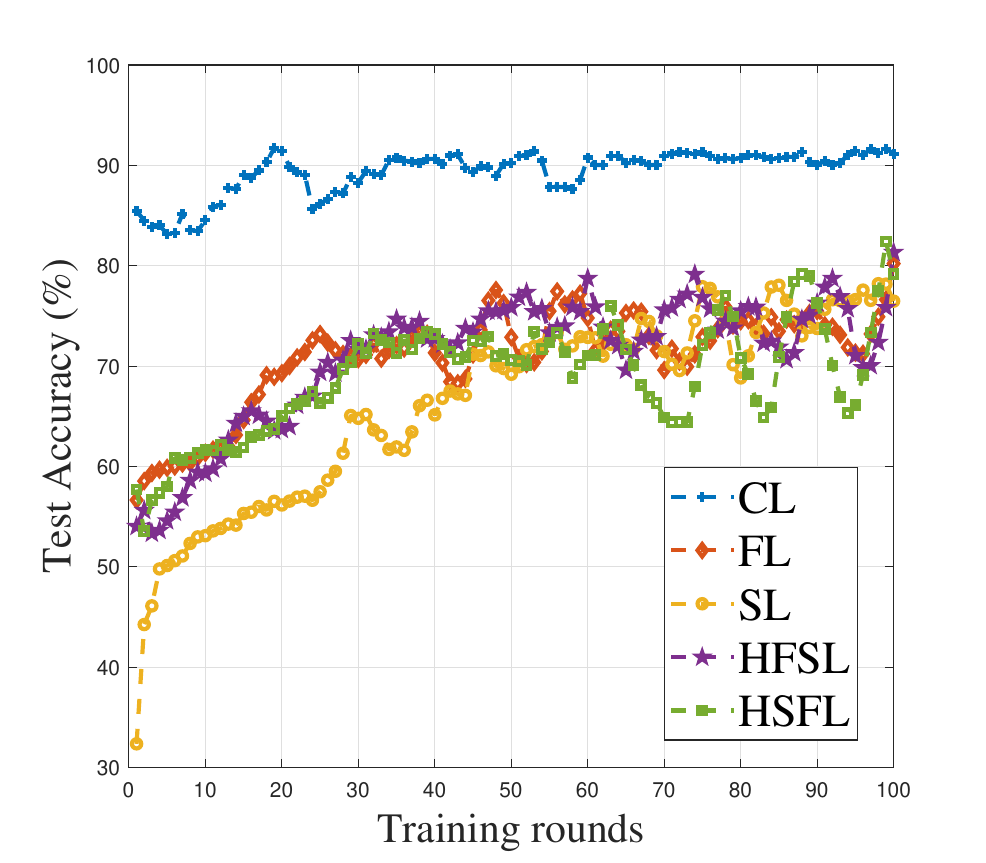}}
    \caption{Test accuracy of CL, FL, SL, HSFL, and HFSL over the training rounds, with four clients upon considering three datasets: a) MNIST; b) Fashion-MNIST; c) MedMNIST.}
    \label{fig3}
\end{figure*}

\begin{table*}[htbp]
\footnotesize
\begin{center}\renewcommand\arraystretch{1.2}
\caption{Performance comparison of different architectures for different data sets in one global round.}
\setlength{\tabcolsep}{0.3mm}{
\begin{tabular}{|ccccccc|}
\hline
\multicolumn{7}{|c|}{\cellcolor{gray!40}\textbf{MNIST}}                                                                                                                                                                                                                                                                                                                                                                                                                                                                                                                  \\ \hline
\multicolumn{1}{|c|}{Architectures} & \multicolumn{1}{c|}{Client network}                                                                                                            & \multicolumn{1}{c|}{Server network}                                                                                                            & \multicolumn{1}{c|}{\begin{tabular}[c]{@{}c@{}}Total comm. cost \\ (MB)\end{tabular}} & \multicolumn{1}{c|}{\begin{tabular}[c]{@{}c@{}}Total comp. cost \\ (FLOPs)\end{tabular}} & \multicolumn{1}{c|}{Test accuracy} & \begin{tabular}[c]{@{}c@{}}Training time\\ (Second)\end{tabular} \\ \hline
\multicolumn{1}{|c|}{CL}            & \multicolumn{1}{c|}{N/A}                                                                                                                       & \multicolumn{1}{c|}{\begin{tabular}[c]{@{}c@{}}cov2d(32,(5,5))+cov2d(64,(3,3))+\\ dense(30976,128)+dense(128,64)+\\ dense(64,10)\end{tabular}} & \multicolumn{1}{c|}{N/A}                                                              & \multicolumn{1}{c|}{N/A}              & \multicolumn{1}{c|}{99.05\%}       & 5.02                                                                  \\ \hline
\multicolumn{1}{|c|}{FL}            & \multicolumn{1}{c|}{\begin{tabular}[c]{@{}c@{}}cov2d(32,(5,5))+cov2d(64,(3,3))+\\ dense(30976,128)+dense(128,64)+\\ dense(64,10)\end{tabular}} & \multicolumn{1}{c|}{N/A}                                                                                                                       & \multicolumn{1}{c|}{60.94}                                                            & \multicolumn{1}{c|}{$1.07*10^8$}      & \multicolumn{1}{c|}{96.83\%}       & 50.83                                                                 \\ \hline
\multicolumn{1}{|c|}{SL}            & \multicolumn{1}{c|}{cov2d(32,(5,5))+cov2d(64,(3,3))}                                                                                           & \multicolumn{1}{c|}{\begin{tabular}[c]{@{}c@{}}dense(30976,128)+dense(128,64)+\\ dense(64,10)\end{tabular}}                                    & \multicolumn{1}{c|}{7090.1}                                                           & \multicolumn{1}{c|}{$7.52*10^7$}      & \multicolumn{1}{c|}{96.20\%}       & 891.02                                \\ \hline
\multicolumn{1}{|c|}{HSFL}          & \multicolumn{1}{c|}{\begin{tabular}[c]{@{}c@{}}cov2d(32,(5,5))+cov2d(64,(3,3))+\\ dense(30976,128)+dense(128,64)+\\ dense(64,10)\end{tabular}} & \multicolumn{1}{c|}{N/A}                                                                                                                       & \multicolumn{1}{c|}{30.47}                                                            & \multicolumn{1}{c|}{$1.07*10^8$}      & \multicolumn{1}{c|}{96.92\%}       &                           49.91                      \\ \hline
\multicolumn{1}{|c|}{HFSL}          & \multicolumn{1}{c|}{cov2d(32,(5,5))+cov2d(64,(3,3))}                                                                                           & \multicolumn{1}{c|}{\begin{tabular}[c]{@{}c@{}}dense(30976,128)+dense(128,64)+\\ dense(64,10)\end{tabular}}                                    & \multicolumn{1}{c|}{7090.4}                                                           & \multicolumn{1}{c|}{$7.52*10^7$}      & \multicolumn{1}{c|}{96.33\%}       &                222.75                        \\ \hline
\multicolumn{7}{|c|}{\cellcolor{gray!40}\textbf{Fashion-MNIST}}                                                                                                                                                                                                                                                                                                                                                                                                                                                                                                          \\ \hline
\multicolumn{1}{|c|}{CL}            & \multicolumn{1}{c|}{N/A}                                                                                                                       & \multicolumn{1}{c|}{\begin{tabular}[c]{@{}c@{}}cov2d(32,(5,5))+cov2d(64,(3,3))+\\ dense(30976,128)+dense(128,64)+\\ dense(64,10)\end{tabular}} & \multicolumn{1}{c|}{N/A}                                                              & \multicolumn{1}{c|}{N/A}              & \multicolumn{1}{c|}{90.34\%}       &      5.02                                                       \\ \hline
\multicolumn{1}{|c|}{FL}            & \multicolumn{1}{c|}{\begin{tabular}[c]{@{}c@{}}cov2d(32,(5,5))+cov2d(64,(3,3))+\\ dense(30976,128)+dense(128,64)+\\ dense(64,10)\end{tabular}} & \multicolumn{1}{c|}{N/A}                                                                                                                       & \multicolumn{1}{c|}{60.94}                                                            & \multicolumn{1}{c|}{$1.07*10^8$}      & \multicolumn{1}{c|}{78.00\%}       &          50.83                                                        \\ \hline
\multicolumn{1}{|c|}{SL}            & \multicolumn{1}{c|}{cov2d(32,(5,5))+cov2d(64,(3,3))}                                                                                           & \multicolumn{1}{c|}{\begin{tabular}[c]{@{}c@{}}dense(30976,128)+dense(128,64)+\\ dense(64,10)\end{tabular}}                                    & \multicolumn{1}{c|}{7090.1}                                                           & \multicolumn{1}{c|}{$7.52*10^7$}      & \multicolumn{1}{c|}{68.51\%}       &                  891.02                                           \\ \hline
\multicolumn{1}{|c|}{HSFL}          & \multicolumn{1}{c|}{\begin{tabular}[c]{@{}c@{}}cov2d(32,(5,5))+cov2d(64,(3,3))+\\ dense(30976,128)+dense(128,64)+\\ dense(64,10)\end{tabular}} & \multicolumn{1}{c|}{N/A}                                                                                                                       & \multicolumn{1}{c|}{30.47}                                                            & \multicolumn{1}{c|}{$1.07*10^8$}      & \multicolumn{1}{c|}{78.26\%}       &                     49.91                                             \\ \hline
\multicolumn{1}{|c|}{HFSL}          & \multicolumn{1}{c|}{cov2d(32,(5,5))+cov2d(64,(3,3))}                                                                                           & \multicolumn{1}{c|}{\begin{tabular}[c]{@{}c@{}}dense(30976,128)+dense(128,64)+\\ dense(64,10)\end{tabular}}                                    & \multicolumn{1}{c|}{7090.4}                                                           & \multicolumn{1}{c|}{$7.52*10^7$}      & \multicolumn{1}{c|}{80.30\%}       &                  222.75                                         \\ \hline
\multicolumn{7}{|c|}{\cellcolor{gray!40}\textbf{MedMNIST}}                                                                                                                                                                                                                                                                                                                                                                                                                                                                                                               \\
\hline
 \multicolumn{1}{|c|}{CL}            & \multicolumn{1}{c|}{N/A}                                                                                                                       & \multicolumn{1}{c|}{ResNet(32)}                                                                                                                & \multicolumn{1}{c|}{N/A}                                                              & \multicolumn{1}{c|}{N/A}              & \multicolumn{1}{c|}{91.73\%}       &       2.03                                                           \\
\hline
\multicolumn{1}{|c|}{FL}            & \multicolumn{1}{c|}{ResNet(32)}                                                                                                                & \multicolumn{1}{c|}{N/A}                                                                                                                       & \multicolumn{1}{c|}{7.18}                                                             & \multicolumn{1}{c|}{$4.27*10^8$}      & \multicolumn{1}{c|}{80.21\%}       &           52.77                                                      \\ \hline
\multicolumn{1}{|c|}{SL}            & \multicolumn{1}{c|}{ResNet(2)}                                                                                                                 & \multicolumn{1}{c|}{ResNet(30)}                                                                                                                & \multicolumn{1}{c|}{1400.02}                                                          & \multicolumn{1}{c|}{$3.2*10^7$}       & \multicolumn{1}{c|}{78.21\%}       &        180.18                                                          \\ \hline
\multicolumn{1}{|c|}{HSFL}          & \multicolumn{1}{c|}{ResNet(32)}                                                                                                                & \multicolumn{1}{c|}{N/A}                                                                                                                       & \multicolumn{1}{c|}{3.59}                                                             & \multicolumn{1}{c|}{$4.27*10^8$}      & \multicolumn{1}{c|}{82.41\%}       &   52.66                        \\ \hline
\multicolumn{1}{|c|}{HFSL}          & \multicolumn{1}{c|}{ResNet(2)}                                                                                                                 & \multicolumn{1}{c|}{ResNet(30)}                                                                                                                & \multicolumn{1}{c|}{1400.18}                                                          & \multicolumn{1}{c|}{$3.2*10^7$}       & \multicolumn{1}{c|}{81.32\%}       &                          45.04                     \\ \hline
\end{tabular}
}
\end{center}
\end{table*}

\section{Open Challenges}
This section discusses open challenges for future research directions of the proposed HSFL and HFSL architectures. Several important examples are discussed as below.

\textbf{Model splitting and resource allocation:} In the context of HSFL, implementing an appropriate splitting scheme, such as determining the value of $M$, can significantly enhance transmission efficiency and reduce client dropout rates. Particularly, in a dynamic D2D network where wireless links among clients are random and ever-changing, the task of defining a suitable value for $M$ becomes a significant challenge, necessitating the development of dynamic model splitting schemes. In the case of HFSL, the structural complexities of the model introduce additional intricacies and challenges to the model splitting problem, especially within D2D networks. For instance, HFSL's model can be divided into multiple subnetworks and allocated to various D2D clients, with both subnetworks and D2D clients represented as two directed graphs. In this scenario, the subnetwork allocation problem in a D2D network is formulated as a subgraph isomorphism problem\cite{Liwang-IOTJ}, which is generally NP-complete and demands solutions with low complexity and high responsiveness. Furthermore, as different model segments (layers) can have varying impacts on training performance, contingent upon each client's local dataset, the model splitting scheme for both HSFL and HFSL should be designed on a client/segment/layer-wise basis, an area that continues to present an open challenge. Additionally, given the heterogeneous resource conditions of clients, optimizing model splitting should involve the joint design of feasible resource allocation strategies, aimed at improving training performance and resource efficiency.

\textbf{Privacy leakage and protection:} Despite the enhanced privacy protection offered by HFSL and HSFL in comparison to SL and FL, privacy breaches remain inevitable even when all participants (i.e., clients or edge servers) are semi-honest (i.e., \textit{honest-but-curious})\cite{Sp-book}. In HSFL, for instance, each participant can receive one or multiple model segments from other participants. As training progresses, each participant eventually gains access to a complete model, exposing HSFL to the same risk of privacy breaches associated with model attacks (e.g., membership inference attacks and model inversion attacks) as FL. In the case of HFSL, although the complete model is not accessible to clients, frequent and direct exchange of data features and gradient information during communication significantly heightens the risk of privacy breaches. While certain existing techniques, such as differential privacy and homomorphic encryption, can mitigate the aforementioned privacy risks to some extent, the associated high computational complexity and potential model performance degradation may not yield satisfactory results. Consequently, designing algorithms with both low computational complexity and guaranteed model accuracy to achieve reduced privacy risks remains a formidable challenge, especially when combining the features of HSFL and HFSL, such as dynamic model partitioning and aggregation strategies.

\textbf{Incentive mechanism design:} While many existing works have examined incentive mechanism designs for FL/SL-based services, the complexity increases when considering HSFL and HFSL. In HSFL/HFSL, the granularity of services (e.g., at the model segment/layer level) provided by each client can be finer, requiring increased cooperation among clients. This introduces challenges like reward transfer among different clients and the secondary distribution of internal rewards within client clusters, which are important areas of concern in HSFL and HFSL.

\textbf{Multiple ML tasks scheduling:} Thanks to the innovative on-board sensors, client devices can collect diverse data types. Moreover, with enhanced multi-core computing processors, a client can engage in multiple ML model training processes simultaneously\cite{MultiFL-TMC}. Consequently, when handling multiple ML training tasks concurrently, it is crucial to select suitable learning architectures (FL, SL, HSFL, or HFSL) and resource scheduling schemes that match the varied requirements of learning tasks and the status of clients, ensuring optimal learning performance. This topic also presents a practical and intriguing research direction.

\textbf{Privacy-oblivious data sharing:} While the fundamental goal of distributed ML is to avoid sharing raw data, exceptions exist where certain clients are permitted to share their training data with trusted peers. For instance, clients may be willing to share photos with family and friends, rather than with strangers, in a privacy-oblivious data-sharing process that can provide more training data. Additionally, clients with limited power supply may offload training data to others to support model training. Data sharing can also lead to client dimensionality reduction in HSFL and HFSL, enabling improved optimizations. To mitigate the risk of privacy disclosure, research is needed on data-sharing strategies, including the measurement of the relationship between privacy disclosure and the amount of shared data and the development of a reputation/credit-based evaluation system.

\section{Preliminary Simulation}
This section evaluates the performance of our proposed hybrid architectures in comparison with centralized learning (CL), FL, and SL on three datasets, namely, MNIST, Fashion-MNIST, and MedMNIST. Each dataset contains multiple classes of different objects, and can thus be utilized to train classification models. Specifically, MNIST includes grayscale images of handwritten digits from `0' to `9'; Fashion-MNIST includes grayscale images of ten different clothing items; MedMNIST includes grayscale images of eight blood cell microscopes. Besides, four clients are supposed to participate in each architecture. To better evaluate the feasibility and stability of the proposed architectures, we adopt a highly non-IID dataset setting \cite{Non-IID-FL} for the clients and datasets. For example, we partition MNIST into four groups according to label spaces, while ensuring that any two groups have different label sets. For example, we assume client-1 has all the training data of class `0' and `1'; client-2 has all the training data of class `2' and `3'; client-3 has all the training data of class `4', `5' and `6'; client-4 has all the training data of class `7', `8' and `9'. Similar dataset settings are applied for Fashion-MNIST and MedMNIST. Besides, Table I demonstrates the model setting for different datasets and architectures. Note that a ResNet(32) model is applied for MedMNIST, where ResNet(2) and ResNet(30) mean that the first two hidden layers and the remaining 30 layers are deployed at the client-side and the server-side, respectively.
%\begin{itemize}
%  \item \textbf{FL:}  Notably, this data setting is adopted for SL, HSFL and HFSL.
%  \item \textbf{SL:} The clients have the same dataset setting as FL. The model we used for MNIST is partitioned and deployed at the client's and edge server's side as shown in Table II.
%  \item \textbf{HSFL:} The clients have the same dataset setting as FL. The model is split into two equal-sized data segments. At each training round, two of the four clients upload the first segment of the model, and the other two upload the second segment of the model for global aggregation.
%  \item \textbf{HFSL:} The clients have the same dataset setting as FL. The model is partitioned and deployed as shown in Table II.
%\end{itemize}

We run simulations with AMD Ryzen 7-5800H@3201MHz as clients and with NVIDIA GeForce RTX3070-8G as an edge server. To better measure the training time, the average uplink/downlink datarate between the edge server and any client is set as 10/50 MB/s, and the average D2D datarate between any two clients is 5 MB/s. In addition, all the clients conduct one local training epoch for each global training round.

Figure 3 demonstrates the test accuracy of different architectures across training rounds for the three datasets, each involving four clients. In general, CL converges rapidly and achieves the highest test accuracy, highlighting its advantages. In contrast, SL converges the slowest, while FL, HSFL, and HFSL achieve relatively close performance on convergence, indicating the feasibility of our proposed architectures. A performance gap between distributed ML and CL is evident, and it widens with the complexity of the dataset and model, primarily due to the non-IID data distribution.
Table I provides a detailed performance comparison of different learning architectures for the three datasets and models. MNIST and Fashion-MNIST exhibit similar performance in most aspects, as they share the same data format, data size, and network model. Compared to FL, HSFL significantly reduces communication costs (by 50\%) and slightly decreases training time while increasing accuracy (e.g., by approximately 2.2\% on MedMNIST). Similarly, compared to SL, HSFL significantly reduces training time (by approximately 75\%) and improves accuracy (e.g., by about 10\% on Fashion-MNIST), although HFSL's communication cost is slightly higher than SL. These results demonstrate that our proposed HSFL and HFSL can significantly reduce communication costs and training time without sacrificing accuracy when compared to conventional FL and SL. Furthermore, a comparison between HSFL and HFSL reveals that HSFL generally has lower communication costs and training time, while HFSL incurs lower computation costs. They achieve similar test accuracy, emphasizing the importance of considering resource availability and performance requirements when implementing HSFL and HFSL.

\section{Conclusion}
In this article, we introduce two hybrid architectures, HSFL and HFSL, for distributed ML in D2D- enabled heterogeneous wireless networks. These architectures combine federated and split approaches. We begin by presenting the basic architectures and conducting performance comparisons to highlight their advantages. We then explore intriguing research directions, pointing out potential challenges and opportunities for future implementations of our proposed architectures. Finally, we perform preliminary simulations to confirm the feasibility of HSFL and HFSL on three datasets, considering highly non-IID data settings.

\bibliographystyle{ieeetr}

\end{document}